\pdfoutput=1

\documentclass[11pt]{article}

\usepackage[final]{acl}

\usepackage{times}
\usepackage{latexsym}

\usepackage[T1]{fontenc}

\usepackage[utf8]{inputenc}
\usepackage{booktabs}

\usepackage{microtype}

\usepackage{inconsolata}

\usepackage{graphicx}
\graphicspath{{pictures/}}

\usepackage{enumitem}
\newlist{inlinelist}{enumerate*}{1}
\setlist[inlinelist,1]{label=(\roman*)}

\usepackage{lipsum}
\usepackage{amsmath}
\usepackage{adjustbox}
\usepackage{linguex}
\usepackage{hyperref}
\usepackage{subcaption}

\title{Is In-Context Learning a Type of Error-Driven Learning? Evidence from the Inverse Frequency Effect in Structural Priming}

\author{Zhenghao Zhou \hspace{.5in} Robert Frank \hspace{.5in} R. Thomas McCoy   \\
    Yale University \\
    370 Temple Street \\
    New Haven, CT 06511 \\
    \texttt{\{herbert.zhou, robert.frank, tom.mccoy\}@yale.edu}}

\begin{document}
\maketitle
\begin{abstract}
Large language models (LLMs) have shown the emergent capability of in-context learning (ICL). One line of research has claimed that ICL is functionally equivalent to gradient descent, a type of error-driven learning mechanism. In this paper, we introduce a new way of diagnosing whether ICL is functionally performing error-driven learning.
Our approach is based on the \textit{inverse frequency effect} (IFE)---a phenomenon in which an agent's behavior is influenced to a greater degree when presented with improbable examples as compared to more likely ones. The IFE has previously been identified in psycholinguistics where humans exhibit the IFE  in the context of structural priming (the tendency for people to produce sentence structures they have encountered recently). In that context, the IFE has been used as evidence that human structural priming must involve error-driven learning mechanisms. In our experiments, we simulated structural priming with ICL and found that LLMs indeed display the IFE, with the effect being stronger in larger models. We conclude that at least in the case we studied, ICL is indeed a type of error-driven learning, supporting the hypothesis that an error signal is implicitly computed in the forward pass during ICL. Our results suggest that both humans and LLMs make use of error-driven processing mechanisms in on-line processing.\footnote{Code is available at: \url{https://github.com/herbert-zhou/ICL_IFE}}

\end{abstract}

\section{Introduction} \label{sec:intro}

To what extent do humans and language models use similar processing mechanisms? 
In some ways, language processing by language models and human learners appears to be substantially different. Humans 
display the 
ability to quickly and flexibly adapt to new contexts, while language models have historically required massive amounts of training data and a large number of parameters to exhibit human-like performance. Yet recent pre-trained large language models (LLMs) have shown the capacity to perform in-context learning (ICL): they flexibly adapt to novel tasks with only a small number of demonstrations provided as prompts in the context window without any parameter updates \citep{brown2020language}. This emergent capability could provide a way to bridge the divide between language models and humans: because ICL enables LLMs---like humans---to flexibly adapt to novel contexts, perhaps ICL shares important properties with the processing mechanisms used by humans.

Within the body of research about why ICL arises and how it operates, one line of work has aimed to deepen our theoretical understanding of ICL by offering functional interpretations of ICL as a kind of implicit gradient descent during inference. \citet{pmlr-v202-von-oswald23a} demonstrate that Transformer models, with appropriate choices of parameters, can process in-context demonstrations in a way that is functionally equivalent to performing gradient updates on the same demonstration examples. \citet{garg2022can}, \citet{zhang2023trained}, and \citet{ahn2024transformers} show that standard Transformers \citep{vaswani2017attention} can be trained to implement learning algorithms for linear regressions under ICL-based training objectives.  \citet{dai2023metaoptimization} provide a mathematical construction showing a dual form between Transformer attention and gradient descent and interpreted ICL as a meta-optimization process that performs implicit fine-tuning. However, \citet{shen2023pretrained} argue that the importance of these theoretical 
demonstrations is limited in that they do not take ICL to be an emergent property, instead assuming a training objective that optimizes for ICL.  These demonstrations therefore deviate from actual LLMs pre-trained with natural data with a language modeling training objective. Indeed, \citeauthor{shen2023pretrained} find inconsistencies between ICL and gradient descent in real models, and therefore leave the equivalence between ICL and gradient descent as an open question.

In this paper, we aim to better characterize \textit{what kind of learning mechanism ICL is} by drawing a connection between ICL and human learning mechanisms, using a case study that allows us to evaluate off-the-shelf LLMs using natural language data. Specifically, we 
investigate
\textbf{whether ICL is a type of \textit{error-driven learning} such that an error signal is implicitly computed during the forward pass}. By focusing on this question, we take some first steps toward understanding whether ICL involves implicit gradient descent because gradient descent is one type of error-driven learning. To approach this question, 
we treat ICL as a processing mechanism of LLMs and borrow insights from methods of studying processing mechanisms in humans: we examine to what extent LLMs show the \textit{inverse frequency effect} (IFE), a phenomenon in human structural priming \citep{branigan2017experimental} that has been argued to  require an error-driven processing mechanism in humans, implicit learning (e.g., \citealp{chang2006becoming}). We focus our attention on the widely studied linguistic phenomenon of the dative alternation where the IFE has been robustly attested in humans. We demonstrate that LLMs not only show a robust IFE under standard fine-tuning (which involves explicit gradient descent, a type of error-driven learning), but also varying degrees of the IFE under the ICL setting, with larger models showing a stronger IFE. Given the correlation between ICL ability and model size, we use our results to conclude that, at least in the cases we study, ICL is indeed an error-driven learning mechanism.

Overall, our results suggest that error-driven learning is an aspect of processing shared by humans and ICL-using LLMs. Our study has implications for both NLP/ML (\ref{ex1}, \ref{ex1.5}, \& \ref{ex2}) and linguistically-motivated analysis of LLMs (\ref{ex3} \& \ref{ex4}):

\ex. We find evidence that ICL in off-the-shelf LLMs can be viewed as a form of error-driven learning in at least some cases.\label{ex1}

\ex. We show how the IFE can serve as a diagnostic for error-driven learning, paving the way for future work that can use this diagnostic to further investigate the circumstances under which ICL functions like error-driven learning.\label{ex1.5}

\ex. We generalize the notion of ICL beyond the standardly assumed prompt format of input-output pairs, establishing a connection between priming and prompting.\label{ex2}

\ex. We show that LLMs qualitatively display an important property of human language processing: the IFE in structural priming.\label{ex3}

\ex. While most human-LLM comparisons focus on representations, our experiments go one step further by analyzing the processing mechanisms used by LLMs.\label{ex4}


\section{Background and Related Work}

In this section, we lay out the building blocks necessary for motivating why we use the IFE to diagnose the error-driven nature of ICL. Our experimental approach is formally described in Section \ref{sec:overview}.

\subsection{Structural Priming in Psycholinguistics} \label{sec:psycholing}

Encountering a syntactic structure can predispose speakers to repeat that structure in the near future, a phenomenon known as syntactic priming
\cite{bock1986syntactic}. For example, after encountering a double object (DO) sentence (e.g., \textit{Alice gave Bob a book}), speakers tend to produce another DO structure (e.g., \textit{The student sent the professor a letter}) rather than a semantically equivalent prepositional dative (PD) structure (e.g., \textit{The student sent a letter to the professor}). Structural priming has been interpreted as an adaptation mechanism, where speakers adapt lexical and syntactic predictions to the current context \cite{jaeger2013alignment}, similar to the way in which LLMs adapt their outputs on the basis of prompts.
Examples of priming typically involve two sentences; we will refer to them as the \textit{prime sentence} and the \textit{target sentence}, where the prime sentence comes first and influences the form of the target sentence.

One important aspect of structural priming is the \textit{inverse frequency effect} \cite{jaeger2008implicit,bernolet2010does,kaschak2011structural}: 
less-preferred syntactic alternatives cause stronger overall priming than more-preferred structures, where the degree to which a structure is preferred is operationalized as its relative frequency in the speaker's experience. 
This can be seen by looking at sentences involving verbs with gradient structural preferences, or \textit{verb biases} (or alternation biases; see \citealp{hawkins2020investigating} for a systematic investigation of verb biases in neural models).  
The verb \textit{give}  allows both prepositional dative (PD)  (\textit{The doctor gave a book to the judge}) and double object (DO) (\textit{The doctor gave the judge a book}) structures, but is biased toward (i.e., occurs more often with) DO in English.    Under the IFE, a ``prime'' sentence involving \textit{give} in its (less preferred) PD structure  will cause a greater priming effect than a DO sentence, i.e., a PD prime will  increase  the probability of a PD occurring more than a DO prime will increase the probability of a DO occurring. The strength of PD priming (i.e., the increase in the probability of a PD target given a PD prime) inversely correlates with the expectation of a PD prime, determined by its verb bias \cite{bernolet2010does}.

Two 
theories have been proposed to account for structural priming. Transient activation theory \cite{pickering1998representation} claims that the activation of structural representations from the prime persists for a short time (in working memory) so that it is easier to reactivate the same structure at the next relevant opportunity. This form of transient activation theory does not however account for the IFE because the amount of residual activation is independent from verb biases and does not involve any error-driven mechanism. In contrast, implicit learning theory \cite{chang2006becoming} claims that humans are continuously and dynamically adjusting their probabilistic knowledge concerning the occurrence of grammatical structures (including verb biases) on the basis of their experience 
and use such information to predict the form of linguistic input. Crucially, under standard theories of learning, the update performed by the learner is error-driven, such that a larger update is performed in situations where the learner's predictions are farther from the truth. In the context of priming, this would mean that priming strength is determined by the difference between the learner's predictions and the actual prime sentence: the less the learner expects the observed prime structure, the larger the resulting error signal is that updates their expectations for that structure in the future, yielding a larger priming effect. Therefore, implicit learning, unlike transient activation, predicts the IFE. The two theories are not mutually exclusive and can co-exist to account for priming, a framing known as the dual mechanism account \cite{tooley2010syntactic}.

In this study, we assume the correctness of the arguments from psycholinguistics 
that 
error-driven learning mechanisms are necessary to explain 
the IFE.
Therefore, by examining whether LLMs show the IFE in the ICL setting, we can infer whether there is an implicit error signal computed in ICL.

\begin{figure*}[htp]
  \centering
  \includegraphics[width=0.95\textwidth]{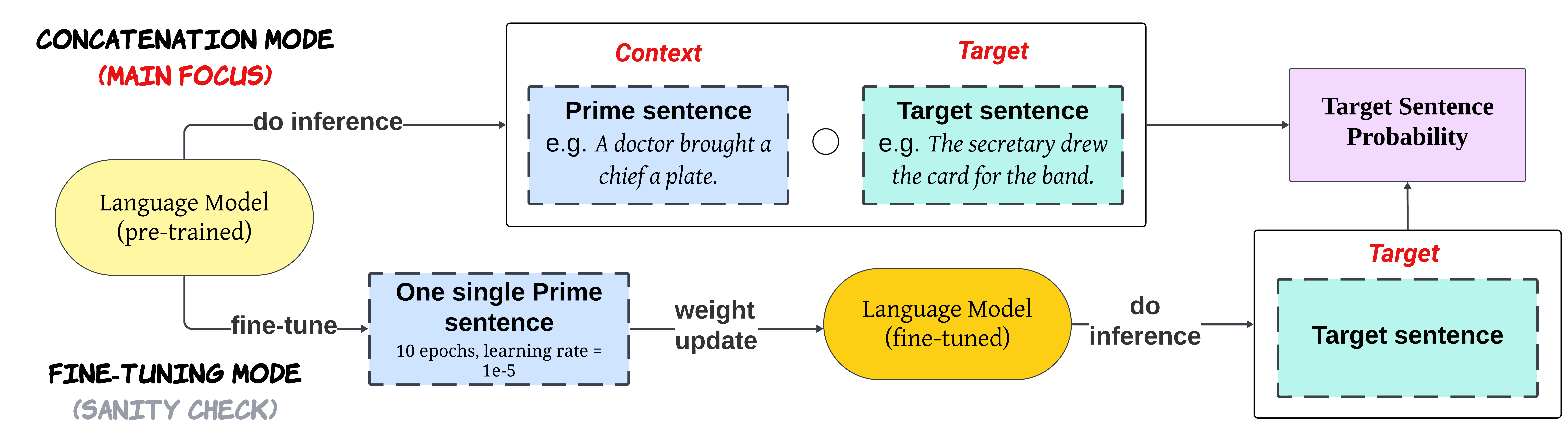}  
  \caption{An overview of our experimental design.}
  \label{overview}
\end{figure*}

\subsection{Structural Priming in Neural Language Models} \label{sec:neural}

Structural priming has been used as a tool for probing the mental representations of structure in humans, under the assumption that these representations
define the notion of similarity underlying priming \citep{branigan2017experimental}. Previous works have adopted this paradigm for probing learned linguistic representations in neural networks.  \citet{van2018neural} and \citet{prasad2019using} show that LSTMs \citep{gulordava2018colorless} are capable of adapting to syntactic structures under fine-tuning.  Specifically, these researchers \textbf{fine-tune} model weights on prime sentences and test target sentence probabilities in the updated model.  This is a direct analog of the implicit learning account of structural priming as it involves weight updates. 
Recently, \citet{sinclair2022structural} have shown that the \textsc{GPT2} family \cite{radford2019language} shows robust structural priming without the use of explicit weight updates; when a prime sentence is  \textbf{concatenated} to its corresponding target sentence, the model assigns a higher probability to the target sentence than it does when the target sentence appears without a prime.
Other works have demonstrated crosslingual structural priming in large language models \cite{michaelov2023structural}, suggesting that structural priming is robustly detected in LLMs. In concurrent work, \citet{jumelet2024language} provide a detailed study of the linguistic factors that give rise to structural priming in current state-of-the-art LLMs, including presenting evidence that LLMs display the IFE; our work differs from theirs in that we study the IFE as a diagnostic for error-driven learning, while \citeauthor{jumelet2024language}'s goal is to compare LLM priming with human priming.  In sum, there is a large body of work demonstrating that LLMs, like humans, show structural priming. This sets the stage for our current study of investigating the processing mechanisms underlying priming. 

\section{Current Study}

\subsection{Overview of Our Approach} \label{sec:overview}

We first clarify our conceptualization of ICL.
In most of the literature, ICL is conceptualized as involving explicit demonstration-answer pairs. However, \citet{chen2024parallel} find that parallel structures in pre-training data substantially contribute to ICL capabilities in LLMs, where parallel structures are defined as pairs of phrases that appear in the same context window and that follow similar templates. Although these parallel structures are not explicitly framed as input-output pairs for a task, they can be conceptualized as in-context examples of less-structured implicit ``tasks'' such as copying n-grams, repeating syntactic constructions, and producing related types of world knowledge. 
Inspired by this perspective, we conceptualize ICL as a more generalized notion that involves \textit{a sensitivity to parallelism} through generic next-token prediction: any text in the context window will influence the model's logits toward preferring similar text. This generalized notion of ICL, namely, the contextual influence of prompts in the context window, is analogous to priming in humans. The fact that \citet{chen2024parallel} showed that parallel structures in pre-training substantially contribute to the emergence of ICL supports the view that we can frame ICL in these more general, parallelism-focused terms.
Under this generalized framing, ICL can produce structural priming: a model that observes a sentence with a particular syntactic structure (say, the DO structure) can use ICL to replicate that syntactic structure, even though the sentence cannot be cleanly viewed as a demonstration-answer pair. That is, the less structured, implicitly defined ``task'' encoded by the DO sentence used as the prompt could be interpreted as ``producing another sentence following the DO structural template exemplified in the prompt.''

Our research question is \textbf{whether an error signal is implicitly computed during the forward pass of processing the prompt in the generalized ICL setting}. We investigate this question by testing whether LLMs show the IFE in the ICL setting.
Given that 
\begin{inlinelist}
    \item it has been argued by psycholinguists that only error-driven learning mechanisms can give rise to the IFE; 
    \item processing the prime sentence in the context window conditions the probability of the target sentence in our generalized notion of ICL; and
    \item standard structural priming in the ICL setting has been robustly observed
\end{inlinelist}, we hypothesize that the strength of the IFE will positively correlate with the strength of the ICL capability of LLMs: the stronger the ICL capability is, the better the error signal will be computed in the forward pass, leading to a stronger IFE. In particular, given recent empirical evidence showing that larger models have stronger ICL capabilities (e.g.,  \citealp{wei2023larger, dong-etal-2024-survey, chen-etal-2025-icleval}), we further hypothesize that the scaling of both model size and data size contributes to LLMs' ICL capability. Thus, we test the hypothesis that \textbf{ICL becomes more like error-driven learning with both model scale and data scale.}

We simulate structural priming across LLMs of various sizes with the two previously explored methods mentioned in Section \ref{sec:neural} (see\
Figure~\ref{overview}). 
The \textbf{\texttt{Fine-Tuning}} mode updates the LLM's parameters on the basis of a single prime sentence, and the updated model is used to infer the probability of the target sentence. The \textbf{\texttt{Concatenation}} mode instead uses (our generalized notion of) ICL, where the prime sentence is  concatenated with the target sentence, so that the prime  constitutes the prompt/context at the point in which the probability of the target sentence is measured. The \texttt{Fine-Tuning} mode serves as a sanity check that LLMs are able to show the IFE with \textbf{explicit} error-driven learning. Because of the nature of gradient descent, changes to model weights during fine-tuning are a function of the degree to which a prime sentence is correctly predicted. A more unexpected prime will therefore give rise to larger gradients and updates, as well as a larger priming effect. Once we have demonstrated this result empirically, the stage will then be set for our main focus: using the \texttt{Concatenation} mode to diagnose whether ICL involves \textbf{implicit} error-driven learning.

\subsection{Corpus} \label{sec:corpus}

We adapted the \textit{Core Dative} \textsc{Prime-LM} Corpus from \citet{sinclair2022structural} to create our dataset. We briefly introduce the relevant properties and refer the readers to the original paper for details. The dative corpus consists of sentences in two forms:

\ex. \textbf{DO}: \texttt{DP$_{subj}$ V DP$_{iobj}$ DP$_{dobj}$} \\
e.g., \textit{\underline{A girl} \underline{bought} \underline{a guy} \underline{a coffee}}. \label{ex5}
    
\ex. \textbf{PD}: \texttt{DP$_{subj}$ V DP$_{dobj}$ Prep DP$_{iobj}$} \\
e.g., \textit{\underline{A girl} \underline{bought} \underline{a coffee} \underline{for} \underline{a guy}}. \label{ex6}

\noindent
Each \texttt{DP} is a determiner with a common noun (120 distinct nouns in total). The corpus was constructed in a way that ensures semantic plausibility and controls for the degree of semantic association and lexical overlap between prime and target sentences.

Since our goal is to study the IFE, we need to ensure variation in the occurrence of verbs with different verb biases in the the prime and target sentences. To do this, for each prime sentence involving one of 22 possible verbs, we sampled 50 target sentences that have no lexical overlap with the prime sentence. This means that the verbs in prime and target sentences are always distinct. Each prime-target pair can be instantiated structurally in 4 different ways, depending on whether the target and prime sentences are PD or DO: $t_{\text{PD}} | p_{\text{PD}}$, $t_{\text{PD}} | p_{\text{DO}}$, $t_{\text{DO}} | p_{\text{PD}}$, $t_{\text{DO}} | p_{\text{DO}}$ (i.e., target sentence $t$ conditioned on prime $p$). This resulted in 92400 prime-target pairs.\footnote{We use $t$ and $p$ for individual target and prime sentences, $T$ and $P$ for sets of targets and primes, and $\mathcal{P}$ for probability.} An example of $t_{\text{PD}} | p_{\text{DO}}$ is \textit{``A doctor brought a chief a plate. The secretary drew the card for the band.''}

We also created an alternative dataset of the same size by replacing each indirect object DP with a pronoun.
This was motivated by a corpus study (see Appendix~\ref{app:pronouns})
that showed that the most common indirect objects in DO sentences are animate pronouns, suggesting that it may be important to consider pronouns in analyses of DO vs.\ PD sentences. The importance of animacy has also previously been noted by \citet{bresnan2007predicting}. The presence and absence of pronouns lead to different verb biases for LLMs, which affect their IFE behaviors. We return to this point in the discussion.

\subsection{Language Models}



We study a set of Transformer models that have been claimed to show ICL capabilities to varying extents \citep{lee2023exploring}. First, we used \textbf{GPT2} \cite{radford2019language} in three of its sizes (\textsc{small, medium, large}), with 85M, 302M, and 708M parameters, respectively. All versions were loaded from the package \texttt{transformerLens} \cite{nanda2022transformerlens}. Second, we used \textbf{Llama2} \cite{touvron2023llama} in three versions: \textsc{7b} (6.5B parameters), \textsc{7b-chat} (6.5B parameters), and \textsc{13b} (13B parameters). All versions were loaded using the Huggingface \texttt{transformers} library \cite{wolf2019huggingface}. Finally, we used \textbf{GPT3-base} \cite{brown2020language} with the \textsc{davinci-002} version (175B parameters), accessed via the OpenAI API. The models vary in size, and correspondingly, in their ICL capabilities, as it has been argued that larger models show stronger ICL capability (e.g., \citealp{wei2023larger, dong-etal-2024-survey, chen-etal-2025-icleval}). We predict that there will be a stronger IFE as size increases.\footnote{We also tested the LSTM models from \citet{gulordava2018colorless} with the \texttt{Concatenation} mode and found no evidence of structural priming.}

\begin{figure}[t]
\centering
  \includegraphics[width=\columnwidth]{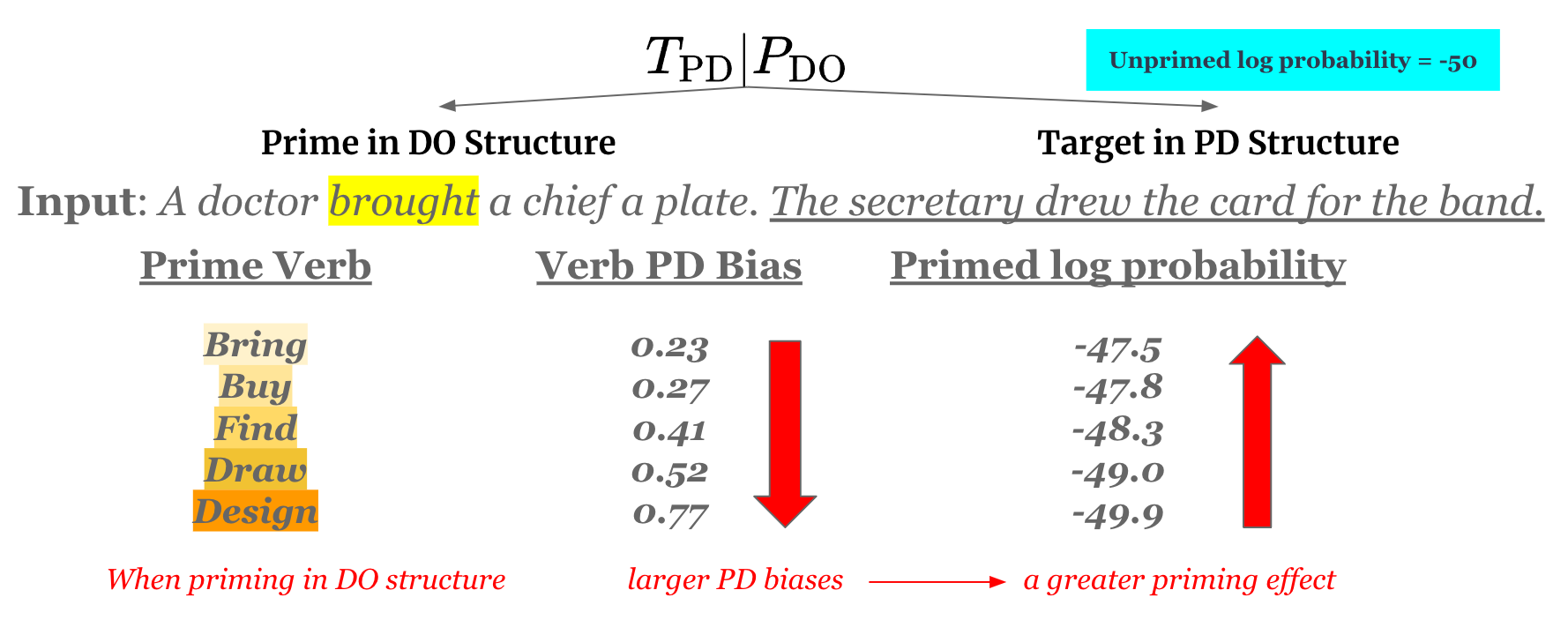}
  \caption{A demonstration of the IFE: a stronger priming effect of a DO prime is predicted as PD-bias increases. The numerical values of the primed log probabilities are hypothetical, for illustration purposes only.} 
  \label{demo}
\end{figure}

\subsection{Quantifying Verb Biases}\label{sec:quantifying_verb_biases}

The verb bias for a specific verb is the likelihood of producing structure $X$ compared to the alternative structure $Y$. In human experiments, baseline verb biases are estimated as the ratio of the number of times  one structure occurs over the sum of both structures; such occurrences can be counted either in natural production experiments or in corpus searches~\cite{zhou2023affects}. Here, we compute a  verb bias for each verb analogously as the ratio of the model probability of one structure over the sum of the probabilities of both structures. The probability of a sentence $s$ is the product of probabilities assigned by LLMs to each token $w_i$: $\mathcal{P}(s) = \prod_i \mathcal{P}(w_i|\overline{w}_{<i})$.
This measures how likely it is for the model to see or produce this sentence. 
Then, given a set of sentences $\mathcal{S}_V$ with ditransitive verb $V$, where each sentence $t_{\text{PD}}$ with structure $\text{PD}$ always has its counterpart $t_{\text{DO}}$ in the opposite structure of $\text{DO}$ (see \ref{ex5} and \ref{ex6}), the $\mathbf{PD}$\textbf{-bias of verb} $\mathbf{V}$ is the mean normalized probability of sentences in structure PD:
\begin{equation} \label{eq:vb}
bias(V, \text{PD}) = \frac{1}{|\mathcal{S}_V|} \sum_{t_{\text{PD}} \in \mathcal{S}_V} \frac{\mathcal{P}(t_{\text{PD}})}{\mathcal{P}(t_{\text{PD}}) + \mathcal{P}(t_{\text{DO}})}
\end{equation}

\noindent
The DO-bias of verb $V,$ i.e., $bias(V, \text{DO})$, is defined analogously. We compute verb bias separately for each model (see Appendix \ref{app:vb} for GPT3), and use a model's biases in our assessment of whether it shows the IFE.  

\subsection{Simulating Structural Priming} \label{sec:simulate}
As stated in Section \ref{sec:overview}, we use two modes to simulate structural priming. Following \citet{van2018neural}, for the \texttt{Fine-Tuning} mode, we update the parameters by fine-tuning the model on a single prime sentence with learning rate $1e^{-5}$ for 10 epochs (see the full fine-tuning details in Appendix \ref{app:ft}), and we use the updated model to do inference on the target sentence. Following \citet{sinclair2022structural}, for the \texttt{Concatenation} mode, we condition a target sentence on a prime sentence through directly concatenating them, separated by a period, without any weight updates. 

The probability of the target sentence after priming is the product of probabilities assigned to its tokens: $\mathcal{P}(t_X | p_X) = \prod_i \mathcal{P}(t_{X_i} | p_X, t_{X_{<i}})$. Under standard priming, the probability of the target sentence $t_X$ should be greater after being primed by a ditransitive prime sentence of either structure: $\mathcal{P}(t_X | p_X) > \mathcal{P}(t_X)$ and $\mathcal{P}(t_X | p_Y) > \mathcal{P}(t_X)$. Being primed by the same structure should induce a larger probability increase than being primed by the opposite structure: $\mathcal{P}(t_X | p_X) > \mathcal{P}(t_X | p_Y)$.

\subsection{Predictions about the Inverse Frequency Effect} \label{section:quant_ife}
Recall that, under the IFE, the priming strength of structure $X$ inversely correlates with the prime verb’s $X$-bias. That is, the degree to which the target production's probability deviates from the baseline is determined by the prime verb. Consider, for example, the case in which the prime sentence has a DO structure and the target sentence has a PD structure (other combinations of structures work analogously).
For each prime verb $V$, we computed the PrimeBias for the PD target structure given a DO prime sentence, where PrimeBias is defined as the normalized target probability primed by this verb over a set of target sentences: 

\begin{equation}\label{eq:PrimeBias}
    \resizebox{0.48\textwidth}{!}{%
    $
    \begin{split}
        PrimeBias(\text{PD} | \text{DO}, V) & = \frac{1}{|T_{\text{PD}}| \cdot |P_{\text{DO}}^V|} \sum_{t_{\text{PD}}\in T_{\text{PD}}} \sum_{p_{\text{DO}}^V\in P_{\text{DO}}^V }  \\ & \frac{ \mathcal{P}(t_{\text{PD}} | p_{\text{DO}}^V)}{\mathcal{P}(t_{\text{DO}} | p_{\text{DO}}^V) + \mathcal{P}(t_{\text{PD}} | p_{\text{DO}}^V)} 
    \end{split} 
    $
}
\end{equation}

\noindent
As shown in Figure~\ref{demo}, the IFE predicts that with a PD target and DO prime sentence, as the prime verb $V$’s PD-bias increases, the prime sentence is less expected, resulting in a larger priming strength towards the DO direction in the target production, i.e., a smaller $PrimeBias(\text{PD} | \text{DO}, V)$ value. Similarly, as PD-bias increases, a PD prime sentence will result in a smaller priming strength towards the PD direction in the target production, i.e., again a smaller $PrimeBias(\text{PD} | \text{PD}, V)$ value. Therefore, when plotting $PrimeBias(\text{PD} | \text{DO}, V)$ and $PrimeBias(\text{PD} | \text{PD}, V)$ against increasing PD verb biases and fitting a line with linear regression, the IFE predicts \textbf{negative slopes for both plots}. Moreover, standard priming predicts that $PrimeBias(\text{PD} | \text{PD}, V)$ should have a higher intercept than $PrimeBias(\text{PD} | \text{DO}, V)$ since the former increases
$\mathcal{P}(T_{\text{PD}})$ more than the latter.\footnote{The other two conditions, namely $T_{\text{DO}} | P_{\text{PD}}$ and $T_{\text{DO}} | P_{\text{DO}}$, are mathematically guaranteed to have slopes that are exactly opposite from $T_{\text{PD}} | P_{\text{PD}}$ and $T_{\text{PD}} | P_{\text{DO}}$, respectively, and the intercept should be 1 minus the intercept of its counterpart.} 

\section{Results and Analysis}

For each model and for each prime verb, we plotted $PrimeBias(\text{PD} | \text{PD}, V)$ and $PrimeBias(\text{PD} | \text{DO}, V)$ against increasing verb biases and used linear regression to find the relationship between priming strength and verb biases. We report the $95\%$ confidence intervals of the fitted slopes to assess the significance of the fitted lines, where intervals below $0$ suggest that the slopes are more likely to be negative.

\begin{figure}[t]
  \centering
  \includegraphics[width=0.9\columnwidth]{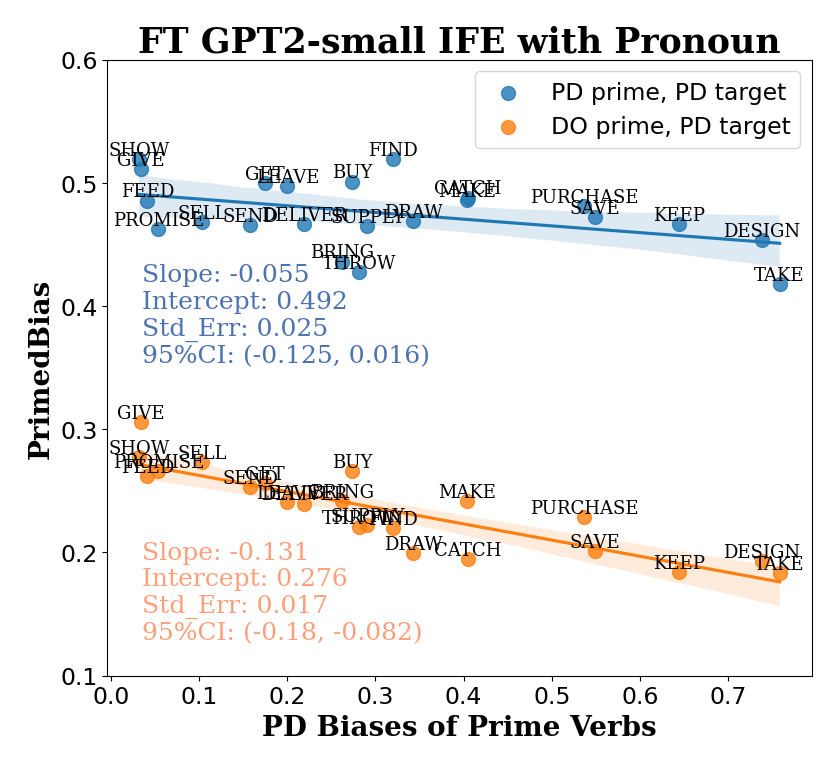}  
  \caption{\textsc{GPT2-small} shows a robust IFE under the \texttt{Fine-Tuning} mode, shown by the two negative slope.}
  \label{ft-mode}
\end{figure}

\begin{figure*}[t]
  \centering
  \includegraphics[width=0.98\textwidth]{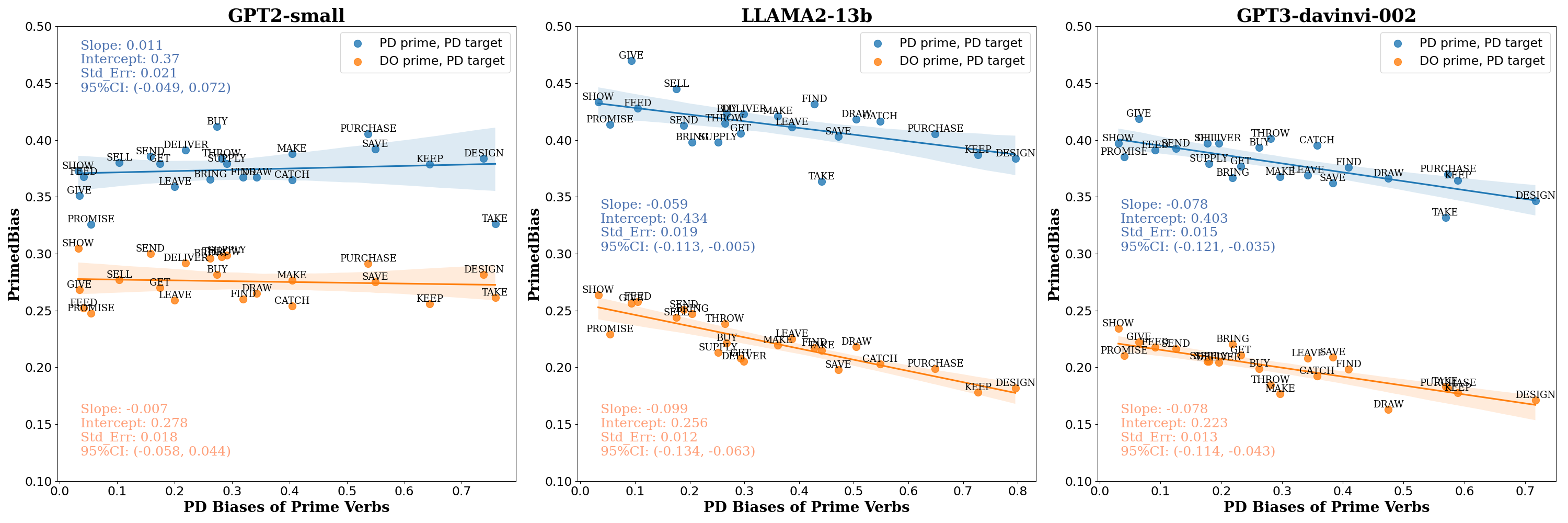}  
  \caption{The IFE across models of different sizes in the \textit{WithPronoun} condition under the \text{Concatenation} mode.}
  \label{ife}
\end{figure*}

\subsection{\texttt{Fine-Tuning} Mode}
We applied the \texttt{Fine-Tuning} mode to \textsc{GPT2-small}.
As is shown in Figure~\ref{ft-mode}, the $T_{\text{PD}} | P_{\text{PD}}$ condition has a larger intercept than the $T_{\text{PD}} | P_{\text{DO}}$ condition, suggesting that the \texttt{Fine-Tuning} mode is able to capture standard structural priming. Further, we observe negative slopes in both lines, suggesting that the \texttt{Fine-Tuning} mode is able to capture the IFE. The $95\%$ confidence interval for the $T_{\text{PD}} | P_{\text{DO}}$ condition is below $0$, demonstrating a stronger IFE than the $T_{\text{PD}} | P_{\text{PD}}$ condition.

Overall, these results show that even the smallest model we analyzed shows the IFE under explicit error-driven weight updates. 
We did not carry out this mode for larger models as fine-tuning on the demonstration with \textsc{GPT2-small} already satisfies its sanity-check purpose: as already noted, we have strong reason to expect that the \texttt{Fine-Tuning} mode will show the IFE, given its explicit gradient-based updates. We now turn to the crucial case of ICL, to see whether the IFE arises even without explicit weight updates.

\subsection{\texttt{Concatenation} Mode} \label{sec:result}

We applied the \texttt{Concatenation} mode to all models. In Figure~\ref{ife}, we show one plot for each of the three types of models (i.e., showing just one variant of that model type and showing only the \textit{WithPronoun} condition) and report the results across all model variants in Table~\ref{tab1} (see Appendix~\ref{app:stats} for the full statistical results). 
For all models across all conditions, the $T_{\text{PD}} | P_{\text{PD}}$ intercept is greater than the $T_{\text{PD}} | P_{\text{DO}}$ intercept, showing the standard structural priming effect, which is consistent with our prediction. For the IFE, we found that all three sizes of \textsc{GPT2} failed to show the IFE, as the slopes are either positive or close to zero. This suggests that, in \textsc{GPT2}, priming strength is not correlated with verb bias under our metric. All three \textsc{Llama2} models showed negative slopes in both cases, which is consistent with the IFE. However, only in the \textit{WithPronoun} $T_{\text{PD}} | P_{\text{DO}}$ condition are the $95\%$ confidence intervals constantly below $0$ across the three models,
suggesting that \textsc{Llama2} displays the IFE but in a noisy way that is not fully robust.
Finally, for \textsc{GPT3}, both $T_{\text{PD}} | P_{\text{PD}}$ and $T_{\text{PD}} | P_{\text{DO}}$ under the  \textit{WithPronoun} condition have their confidence intervals below $0$, while the confidence intervals in both case in the \textit{NoPronoun} condition contain $0$.

The \textit{WithPronoun} DO-PD conditions of all three \textsc{Llama2} models and \textsc{GPT3} are statistically significant with $p < 0.0001$. Models at least as large as \textsc{Llama2-13b} show at least marginal statistical significance in all conditions while smaller models do not. These results support our hypothesis that larger models show a more significant IFE.

Therefore, besides corroborating the finding from prior work that LLMs show structural priming effects, the current results suggest that \textbf{larger models tend to show a stronger IFE than smaller models, meaning that IFE strength correlates with ICL capability} since larger models also typically perform better at ICL than smaller models \citep{brown2020language}. Assuming that ICL capability correlates with LLM size, given the currently observed pattern, we predict that larger models such as \textsc{GPT4} should show an even stronger IFE.

\begin{table*}
    \centering
    \begin{adjustbox}{width=0.9\textwidth}
\begin{tabular}{llllllll}
    \toprule
                &   &  \multicolumn{2}{c}{Intercept}  &  \multicolumn{2}{c}{Slope} &  \multicolumn{2}{c}{$95\%$ CI of Slopes} \\ \cmidrule(lr){3-4} \cmidrule(lr){5-6} \cmidrule(lr){7-8}
    
                Model  &  Pronoun Obj? & PD-PD & DO-PD & PD-PD & DO-PD & PD-PD & DO-PD \\
    \midrule
          GPT2-small &     True &            0.370 &            0.278 &    0.011 &   -0.007 &  (-0.049, 0.072) &  (-0.058, 0.044) \\
          GPT2-small &    False &            0.746 &            0.653 &    0.014 &    0.006 &   (-0.042, 0.07) &   (-0.06, 0.072) \\
         GPT2-medium &     True &            0.351 &            0.256 &   -0.013 &   -0.026 &  (-0.078, 0.053) &  (-0.073, 0.022) \\
         GPT2-medium &    False &            0.748 &            0.590 &   -0.023 &   -0.035 &  (-0.079, 0.032) &  (-0.125, 0.054) \\
          GPT2-large &     True &            0.330 &            0.241 &    0.011 &   -0.037$^{+}$ &  (-0.043, 0.065) &   (-0.09, 0.015) \\
          GPT2-large &    False &            0.698 &            0.487 &   -0.003  &    -0.02 &  (-0.062, 0.055) &  (-0.098, 0.058) \\
            Llama-7b &     True &            0.392 &            0.229 &    -0.02  &   -0.086$^{****}$ &  (-0.067, 0.026) & \textbf{(-0.126, -0.045)} \\
            Llama-7b &    False &            0.807 &            0.627 &   -0.026  &   -0.111$^{+}$ &  (-0.102, 0.049) &  (-0.279, 0.057) \\
       Llama-7b-chat &     True &            0.413 &            0.263 &   -0.012  &   -0.095$^{****}$ &  (-0.067, 0.043) & \textbf{(-0.146, -0.044)} \\
       Llama-7b-chat &    False &            0.788 &            0.605 &   -0.013  &   -0.102  &  (-0.115, 0.089) &  (-0.289, 0.085) \\
           Llama-13b &     True &            0.434 &            0.256 &   -0.059$^{**}$ &   -0.099$^{****}$ & \textbf{(-0.113, -0.005)} & \textbf{(-0.134, -0.063)} \\
           Llama-13b &    False &            0.859 &            0.685 &   -0.066$^{+}$ &   -0.177$^{*}$ &  (-0.163, 0.031) & (-0.388, 0.033) \\
    GPT3-davinci-002 &     True &            0.403 &            0.223 & -0.078$^{****}$ & -0.078$^{****}$ & \textbf{(-0.121, -0.035)} & \textbf{(-0.114, -0.043)} \\
    GPT3-davinci-002 &    False &            0.851 &            0.632 &   -0.064$^{+}$ &   -0.145$^{*}$ &  (-0.153, 0.025) &  (-0.301, 0.012) \\
    \bottomrule
\end{tabular}
    \end{adjustbox}
    \caption{The intercept, slope, and $95\%$ confidence interval of the fitted lines for each condition under the \texttt{Concatenation} mode. We mark the $p$-values of the fitted slopes with the following notation: * for $p \leq 0.05$, ** for $p \leq 0.01$, **** for $p \leq 0.0001$, and $+$ for $p \leq 0.1$ (marginally significant). $95\%$ confidence intervals of the fitted slopes below zero are bold (which means the fitted slopes are more statistically significant to be negative).}
    \label{tab1}
\end{table*}

\subsection{The Distinction between the \textit{WithPronoun} vs.\ \textit{NoPronoun} Conditions}

As shown in Table~\ref{tab1}, the majority of cases with their $95\%$ confidence intervals excluding $0$ are the \textit{WithPronoun} $T_{\text{PD}} | P_{\text{DO}}$ cases. Why the IFE arises more strongly in the \textit{WithPronoun} condition than the \textit{NoPronoun} condition remains unclear. One major difference between these conditions lies in the default verb biases: as shown in Figure~\ref{bias} in Appendix~\ref{app:vb}, \textsc{GPT3} shows an overwhelming bias towards PD without pronouns but a reverse pattern favoring DO with pronouns. This pattern holds across all models and is consistent with statistics obtained from our corpus parse. One possible conclusion is that the most common indirect object DPs in the DO sentences are animate pronouns, causing the model to assign a higher probability to pronoun sentences. 
The asymmetry between the \textit{WithPronoun} and \textit{NoPronoun} conditions can be explained by our claim that \textbf{ICL becomes more like error-driven learning with both model scale and data scale} (see Section~\ref{sec:overview}). The role of model scale has already been demonstrated in Table~\ref{tab1} (i.e., larger models show stronger IFE in general); here we elaborate on the role  of data scale, which is somewhat more subtle.

In the corpus study mentioned in Appendix~\ref{app:pronouns}, we found that pronominal indirect object (IO) sentences are more common than non-pronominal ones in the OpenWebText corpus. 
Since larger models have seen more pronominal IO sentences compared to non-pronominal ones, they are expected to perform more like error-driven learning for pronominal IO sentences than non-pronominal IO sentences in ICL, under the assumption that data scale (like model scale) increases the strength of the IFE. Such relationships between data scale and the strength of ICL-related effects have also been observed in \citet{mccoy2024embers}, who showed that LLMs displayed better ICL performance on high-probability sentences than low-probability ones. This observation could also potentially explain the interesting finding from \citet{sinclair2022structural} that structural priming in LLMs (to a greater degree than in humans) is modulated by semantic plausibility: semantically plausible inputs are better represented in the training data, leading to more ICL/structural priming.

One further empirical observation from our experiments supports this reasoning: for Llama2-13B and GPT3, we observed (marginally) significant IFE (p < 0.1 or p < 0.05) for the \textit{NoPronoun} conditions, although it is a smaller effect than in the \textit{WithPronoun} conditions. The fact that there are significant effects in both cases suggests that the asymmetry between the two conditions is quantitative rather than qualitative. This graded view aligns well with the data-scale-based account: in addition to the difference of IFE significance explained by model scale, the across-condition differences within single models are accounted for by the fact that the model is trained on more with-pronoun dative sentences than without-pronoun ones, which makes the model better approximate error-driven learning in ICL with the with-pronoun ones, which leads to a more significant IFE as we have observed.

\section{Discussion and Conclusion}

\paragraph{Evidence that ICL is implicitly an error-driven learning mechanism} 
We started with the question of whether ICL could be a processing mechanism of LLMs that can capture the flexibility and adaptability of human learning mechanisms. One type of error-driven learning mechanism that supports flexible adaptation to data is gradient descent.  Motivated by previous theoretical proposals that ICL could in principle be implicitly performing gradient descent or fine-tuning, we attempted to better characterize what kind of learning ICL is in actual models.  We focused on one particular aspect of ICL: whether it involves an implicit error signal during the forward computation. In contrast to previous analyses of ICL, we test standard LLMs with natural language data. 

Having established a connection between ICL and human structural priming, we used the IFE to diagnose whether LLMs behave as if they are implicitly performing error-driven learning when processing a prime sentence. The very nature of the IFE requires an error-driven mechanism: processing must in some way be sensitive to the error signal, i.e., the amount of divergence from a model’s expectations. In the case of DO/PD sentences, the IFE is defined based on verb biases: the error signal is created by the degree of mismatch between the expectation on structural alternatives based on the verb bias of the prime verb and the actual perceived prime structure. The psycholinguistic literature discussed in Section~\ref{sec:psycholing} was the first to identify the logic of this connection between the IFE and error-driven learning, and it is this logic that we are employing in the framing of this paper. Note however that our argumentation does not in any way assume that ICL involves the same mechanism as human implicit learning.

We found that LLMs do indeed display the IFE in many cases, particularly in larger models. These findings support the hypothesis that an error signal is implicitly computed in the forward pass of ICL. Differences between models suggests that ICL only takes on a gradient-descent-like nature in larger models. 
More speculatively, these results raise the possibility that error-driven learning might be a crucial property that enables generalization from a small number of samples, an ability that is shared by human learners and (to some extent) LLMs performing ICL. 
Our study not only provides behavioral results that align LLM and human behavior in structural priming at the processing mechanism level, but also demonstrates the possibility of studying the nature of ICL with off-the-shelf pre-trained LLMs and with naturalistic data.

\paragraph{ICL as a consequence of language modeling} ICL is typically understood as involving demonstration-answer pairs in the prompt. Inspired by data-centric views that explain ICL from the distributional properties of pre-training data
, we proposed a generalized notion of ICL that is sensitive to general parallelisms.
Therefore, ICL could be viewed as a side product of the general propensity for structural parallelism in the language modeling task. We leave this perspective for future study.

\paragraph{Future Directions} \label{sec:future} In this study, we focused on one single ``task'', namely selecting DO versus PD sentence structures. If our reasoning is correct, such that it is indeed the implicit error signal of the ICL that results in LLMs' capability of capturing the IFE, then we predict that the IFE diagnostics could be generalized to other ICL tasks, even non-linguistic tasks. Future work could extend our current method to additional ICL tasks. Finding the IFE on a wider range of tasks would strengthen the claim that ICL is driven by implicit error-driven computation, while not observing the IFE on other tasks would indicate that ICL only sometimes displays a gradient-descent-like character.


\section*{Limitations}

\paragraph{Behavioral versus Mechanistic Accounts} Although ICL is generally identified as a phenomenon at the behavioral level, having an explanation at the mechanistic level would bring greater interpretability and could contribute to more concrete theory building. Our current study, despite using real pre-trained models and naturalistic data, remains at the behavioral level and is empirical in nature. Given our current contribution of establishing a connection between ICL and human priming and using the IFE as a way to diagnose whether ICL works in a gradient-descent-like manner or not,  future work could improve our understanding by incorporating techniques from mechanistic interpretability to explain our current finding at the mechanistic level.

\paragraph{Examining the IFE in Other Models} As ICL capability is often argued to scale with model size, we predict in Section \ref{sec:result} that the IFE effect will be more robust in larger models. Although the difference in the IFE behavior between \textsc{GPT2} and \textsc{GPT3-base} is substantial, we do not believe that we have observed a saturation of the IFE because increasing the confidence level to $99\%$ could still result in confidence intervals containing $0$. \textsc{GPT3-base} is currently the biggest model on which we have access to the logit predictions, but we believe the same behavioral test could be applied to larger models in order to verify our prediction.

\section*{Acknowledgments}
We would like to thank the anonymous reviewers from the ARR June and October 2024 cycles, ESSLLI Student Session 2024, and AMLaP 2024 for their thoughtful comments, which helped us refine this paper. We would also like to thank the members of the Computational Linguistics at Yale Lab, the Yale Computation and Cognition Joint Lab, and the Yale Linguistics Department for helpful feedback. We thank the Yale Center for Research Computing for guidance and use of the research computing infrastructure, specifically the Grace cluster. Any errors are our own.

\bibliography{priming}

\appendix

\section{Finding pronoun probabilities in the OpenWebText corpus with \texttt{spaCy}} \label{app:pronouns}
As is mentioned in Section \ref{sec:corpus}, in order to estimate the verb biases represented in the training corpus of GPT2 models, we parsed a fragment (approximately 160 million tokens) of the OpenWebText corpus \citep{Gokaslan2019OpenWeb} with the Python package \texttt{spaCy} \citep{spaCy} to get an estimate of the DO vs.\ PD ratio for each verb. Specifically, we used the \texttt{en\_core\_web\_trf} specification of the \texttt{spaCy} model, and we identified the set of dative alternation sentences by doing dependency parsing on each sentence. We found that the verb biases from the corpus are less well-represented in GPT2 models, motivating our decision to estimate verb biases by using model judgments rather than corpus statistics; see Section~\ref{sec:quantifying_verb_biases}.

The corpus parse results motivated us to investigate the impact of whether the indirect object is a pronoun on the verb biases. Thus, we constructed the \textit{WithPronoun} version of the corpus. To do this, we approximated the distribution of the natural occurrence frequencies over the set of English pronouns in dative alternation sentences from a fragment of the OpenWebText corpus \citep{Gokaslan2019OpenWeb}, which is a reasonable proxy for the closed WebText corpus that was used to train the \textsc{GPT2} models. Then, we counted the frequencies of the set of English pronouns that occurred as the indirect object of the ditransitive verb. The list of pronouns and their frequencies are presented in Table \ref{tab2}, sorted by frequency.

\begin{table}[h!] 
\centering
 \begin{tabular}{||c c ||} 
 \hline
 Pronoun & Frequency  \\  [0.5ex] 
 \hline\hline
  you & 4621 \\ 
  me & 2962  \\
  us & 2959  \\
  him & 2210  \\
  them & 1847  \\ 
  it & 1297  \\ 
  her & 738  \\  [0.5ex] 
 \hline
 \end{tabular}
 \caption{The respective frequencies of the English pronouns occurring as the indirect object of ditransitive sentences in a fragment of the OpenWebText corpus.}
 \label{tab2}
\end{table}

To convert the existing dative alternation priming corpus to the \textit{WithPronoun} version, we replaced the indirect object of every sentence in the existing corpus with one of the pronouns through random sampling according to their respective relative frequencies.

\section{Fine-tuning details} \label{app:ft}

As is presented in Section \ref{sec:simulate}, to simulate structural priming in the \texttt{Fine-tuning} mode, we fine-tuned a pre-trained \textsc{GPT2-small} model on every prime sentence and used the updated model to do inference on the target sentences.

We loaded the pre-trained \textsc{GPT2-small} model from the \texttt{TransformerLens} \citep{nanda2022transformerlens} package and used the \texttt{train} function from \texttt{TransformerLens} to do fine-tuning. To avoid catastrophic forgetting during fine-tuning, we applied a regularization term to the loss function for gradient descent. We randomly sampled a fixed set of 5000 adjacent tokens from the OpenWebText (so that it resembles the distribution of the pre-training data) and computed the loss on them of the pre-trained \textsc{GPT2-small} model. Then, at each step during fine-tuning, we added to the loss term the squared difference between the current loss and the raw (pre-trained) loss of the model on these 5000 tokens, scaled by a coefficient $\lambda = 0.8$. We found that this regularization term helped keep the model stable during fine-tuning on a single sentence.

\begin{table}[h] 
\centering
 \begin{tabular}{||c c ||} 
 \hline
 Parameter & Value  \\ [0.5ex] 
 \hline\hline
  number of epochs & $10$ \\ 
  batch size & $1$  \\
  learning rate & $1e^{-5}$  \\
  optimizer & AdamW  \\
  lambda & 0.8  \\ [0.5ex] 
 \hline
 \end{tabular}
 \caption{Hyperparameters used as the training configuration for the \texttt{Fine-tuning} mode of structural priming on \textsc{GPT2-small}.}
 \label{tab3}
\end{table}

We did a hyperparameter search and chose the set of parameters in Table \ref{tab3}. We used the default values from \texttt{TransformerLens} for the rest of the relevant hyperparameters (such as \texttt{warmup}, \texttt{maximum gradient norm}, etc.).

\section{Verb Biases represented in LLMs} \label{app:vb}

\begin{figure*}[h]
    \centering
    \begin{subfigure}[b]{0.48\textwidth}
        \centering
        \includegraphics[width=\textwidth]{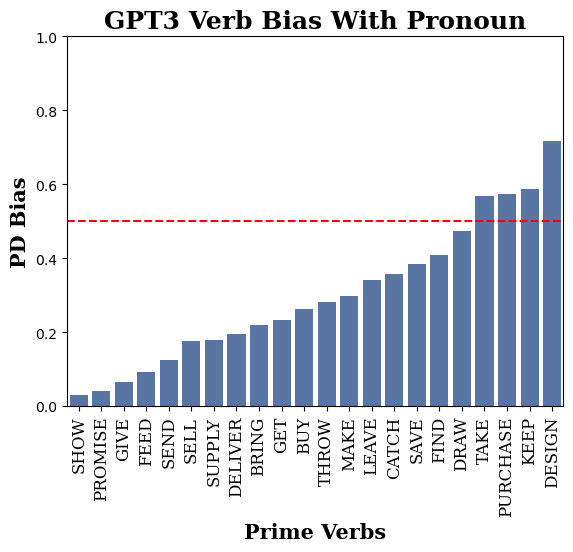}
    \end{subfigure}
    \hfill
    \begin{subfigure}[b]{0.48\textwidth}
        \centering
        \includegraphics[width=\textwidth]{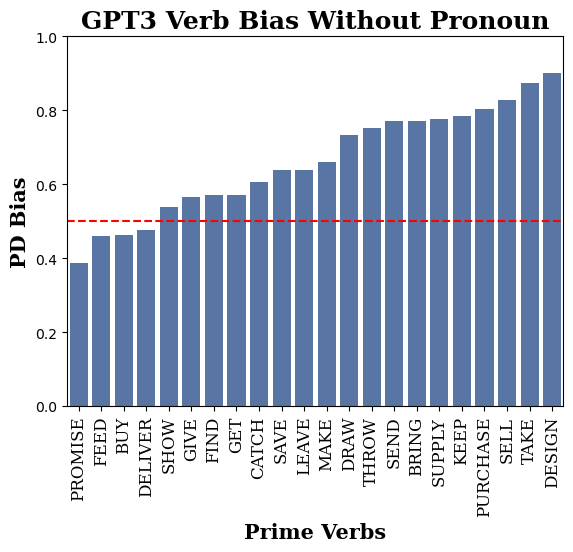}
    \end{subfigure}
    \caption{Comparison of PD biases with (left) and without (right) pronoun for \textsc{GPT3}. As is shown in Equation.~\ref{eq:vb}, a high PD-bias means a larger proportion of probability assigned to the PD structure against the DO structure in LLMs.}
    \label{bias}
\end{figure*}

See Figure~\ref{bias} for details of the verb biases in GPT3.


\section{Full Statistics of the IFE Experiments in \texttt{Concatenation} Mode} \label{app:stats}

See Table~\ref{tab_full} (as a supplement to Table~\ref{tab1}) for the full statistical metrics of the IFE experiments reported in Section~\ref{sec:result}.
 
\begin{table*}
    \centering
    \begin{adjustbox}{width=\textwidth}
\begin{tabular}{cccccccccccc}
\toprule
            &   &  \multicolumn{2}{c}{Std\_Err}  &  \multicolumn{2}{c}{$r$-value} &  \multicolumn{2}{c}{$p$-value}  &  \multicolumn{2}{c}{$R^2$}  &  \multicolumn{2}{c}{RMSE} \\ \cmidrule(lr){3-4} \cmidrule(lr){5-6} \cmidrule(lr){7-8} \cmidrule(lr){9-10} \cmidrule(lr){11-12}

            Model  &  Pronoun Obj? & PD-PD & DO-PD & PD-PD & DO-PD & PD-PD & DO-PD & PD-PD & DO-PD & PD-PD & DO-PD \\
\midrule
      GPT2-small &     True &          0.021 &          0.018 &          0.119 &         -0.088 &          0.597 &          0.698 &     0.014 &     0.008 &       0.020 &       0.017 \\
      GPT2-small &    False &          0.020 &          0.023 &          0.154 &          0.058 &          0.495 &          0.798 &     0.024 &     0.003 &       0.016 &       0.019 \\
     GPT2-medium &     True &          0.023 &          0.017 &         -0.123 &         -0.327 &          0.585 &          0.138 &     0.015 &     0.107 &       0.023 &       0.016 \\
     GPT2-medium &    False &          0.019 &          0.031 &         -0.258 &         -0.245 &          0.247 &          0.273 &     0.067 &     0.060 &       0.017 &       0.027 \\
      GPT2-large &     True &          0.019 &          0.018 &          0.129 &         -0.416 &          0.568 &          0.054 &     0.017 &     0.173 &       0.019 &       0.018 \\
      GPT2-large &    False &          0.020 &          0.027 &         -0.038 &         -0.163 &          0.868 &          0.469 &     0.001 &     0.026 &       0.018 &       0.024 \\
        Llama-7b &     True &          0.016 &          0.014 &         -0.270 &         -0.803 &          0.224 &          0.000 &     0.073 &     0.645 &       0.015 &       0.013 \\
        Llama-7b &    False &          0.027 &          0.059 &         -0.215 &         -0.387 &          0.336 &          0.076 &     0.046 &     0.149 &       0.019 &       0.042 \\
   Llama-7b-chat &     True &          0.019 &          0.018 &         -0.138 &         -0.766 &          0.541 &          0.000 &     0.019 &     0.587 &       0.018 &       0.017 \\
   Llama-7b-chat &    False &          0.036 &          0.066 &         -0.082 &         -0.327 &          0.718 &          0.137 &     0.007 &     0.107 &       0.024 &       0.044 \\
       Llama-13b &     True &          0.019 &          0.012 &         -0.568 &         -0.872 &          0.006 &          0.000 &     0.323 &     0.760 &       0.018 &       0.011 \\
       Llama-13b &    False &          0.034 &          0.074 &         -0.400 &         -0.473 &          0.065 &          0.026 &     0.160 &     0.224 &       0.019 &       0.042 \\
GPT3-davinci-002 &     True &          0.015 &          0.013 &         -0.755 &         -0.813 &          0.000 &          0.000 &     0.570 &     0.662 &       0.013 &       0.011 \\
GPT3-davinci-002 &    False &          0.031 &          0.055 &         -0.415 &         -0.507 &          0.055 &          0.016 &     0.172 &     0.257 &       0.020 &       0.035 \\
\bottomrule
\end{tabular}
    \end{adjustbox}
    \caption{The standard error, $r$-value, $p$-value, $R^2$ coefficient, and RMSE score for each condition under the \texttt{Concatenation} mode.}
    \label{tab_full}
\end{table*}

\end{document}